\documentclass{article}

\usepackage{arxiv}

\usepackage[utf8]{inputenc} 
\usepackage[T1]{fontenc}    
\usepackage{hyperref}       
\usepackage{url}            
\usepackage{booktabs}       
\usepackage{amsfonts}       
\usepackage{nicefrac}       
\usepackage{microtype}      
\usepackage{lipsum}
\usepackage{algpseudocode}
\usepackage{graphicx}
\usepackage{siunitx}
\usepackage{xcolor}
\usepackage{dcolumn}
\usepackage{array}
\usepackage{multirow}
\usepackage{subcaption}

\title{Zero-shot topic generation}

\author{
 Oleg Vasilyev, Kathryn Evans, Anna Venancio-Marques, and John Bohannon \\
  Primer\\
  San Francisco, California\\
  \texttt{\{oleg,kathryn,anna,john\}@primer.ai} \\
}

\begin{document}
\maketitle
\begin{abstract}
We present an approach to generating topics using a model trained only for document title generation, with zero examples of topics given during training. We leverage features that capture the relevance of a candidate span in a document for the generation of a title for that document. The output is a weighted collection of the phrases that are most relevant for describing the document and distinguishing it within a corpus, without requiring access to the rest of the corpus. We conducted a double-blind trial in which human annotators scored the quality of our machine-generated topics along with original human-written topics associated with news articles from The Guardian and The Huffington Post. The results show that our zero-shot model generates topic labels for news documents that are on average equal to or higher quality than those written by humans, as judged by humans.
\end{abstract}

\section{Introduction}
The novel method that we present belongs to an overlap of summarization and categorization tasks traditionally described as keyphrase generation, keyphrase extraction, and topic modeling \cite{eirini2019areview, erion2019survey, erion2019keyphrase, chang2009reading}. 

Topics have multiple purposes. One is to provide a succinct summary of a document by listing its core concepts \cite{erion2019keyphrase, meng2017deepkeyphrase}. This is a document-centric version of topics that enables a reader to quickly grasp the most important themes within a text without  reading it. On the other end of the spectrum, topics also serve the purpose of describing how groups of documents relate to each other and the distribution of themes across an entire corpus. This is essentially a categorization task, where the category labels may be predetermined — such as the various sections of a newspaper — or generated \textit{de novo} from the corpus itself \cite{erion2019survey, david2012probabilistic}. 

This reveals two challenges worth noting. First, there is ambiguity around the term "topic" itself. Depending on the version of the task, a "topic" may refer to a string used to describe and label a single document or a cluster of multiple documents, or "topic" may refer to the document cluster itself. The method we describe in this paper serves primarily to generate labels for single documents. (Of course, the generated single-document topic labels, or embedding representations thereof, can serve as the basis for downstream document clustering.)

A second challenge worth noting is that the different versions of topic modeling conflict in their requirements. For example, for the task of summarizing the key themes of a single document, the optimum topic labels may call for highly specific keyphrases. But for the task of describing how that single document relates thematically to the rest of a corpus, more generic topic labels may be required. Due to the contrasting nature of these objectives, we put emphasis here on generating topic labels that are optimal for summarizing a single document.

A method for efficiently generating highly relevant topic labels for documents has many advantages. Traditional topic modeling methods that approach topics from within a generic theme, such as those based on LDA \cite{david2003latent} or clustering, have often been found to be too general or vague to provide a good summary \cite{eirini2019areview}.  Additionally, identifying higher level themes and document groupings typically requires computation on the entire corpus to assign topics, which is ill-suited to streaming contexts \cite{akash2017autoencoding, Jason2019prediction}.

Traditional topic label extraction methods can be split into two steps: First, the creation of a candidate list of potential topics, and second, the selection of representative topics from the candidates by ranking or filtering \cite{hasan2014survey, erion2019survey}. The candidate lists can be generated through noun phrase extraction or other part of speech patterns \cite{liu2010topic, meng2017deepkeyphrase}. However, many of the ranking techniques are supervised and built on top of features based on TF-IDF or TopicRank \cite{bougouin2013topicrank}. These methods only leverage basic word statistics. As documents often have frequently occurring terms with don't relate to the task \cite{Jason2019prediction}, effective ranking requires a representation of a term's semantic meaning and significance to the document \cite{meng2017deepkeyphrase}.  

We introduce a novel topic generation method that utilizes a universal language model trained for document summarization \cite{oleg2019headline}.
This model selects candidate spans as it goes down a text generation path. Rejected candidate spans that lay near the generation path still have high significance to the document and thus provide valuable information for summarization. Though these second-rank candidate spans did not make the cut for the final summary output, they turn out to be valuable as topic prototypes. Thus a model trained only for the task of document summarization can be used as is for topic generation, with no need for topic-specific training.
 
Following the traditional extraction pattern, our method begins by extracting all noun phrases from a document as candidate spans, and then filters them by overlap with topics prototypes obtained by our summarization model. The resulting set of spans is a mixture of keyphrases suitable for summarizing the document as well as more generic concepts suitable for thematically grouping the document with others in a corpus.

To assess the quality of our machine-generated topic labels, we assembled a corpus of documents from the news websites of The Guardian and The Huffington Post, both of which include topic labels. We conducted a double-blind trial in which human annotators were presented with the articles and topic labels (either machine-generated or the original human-written labels) and asked to score their quality.

\section{Methods}

\subsection{Our strategy}

\begin{figure}
    \centering
    \includegraphics[width=14cm]{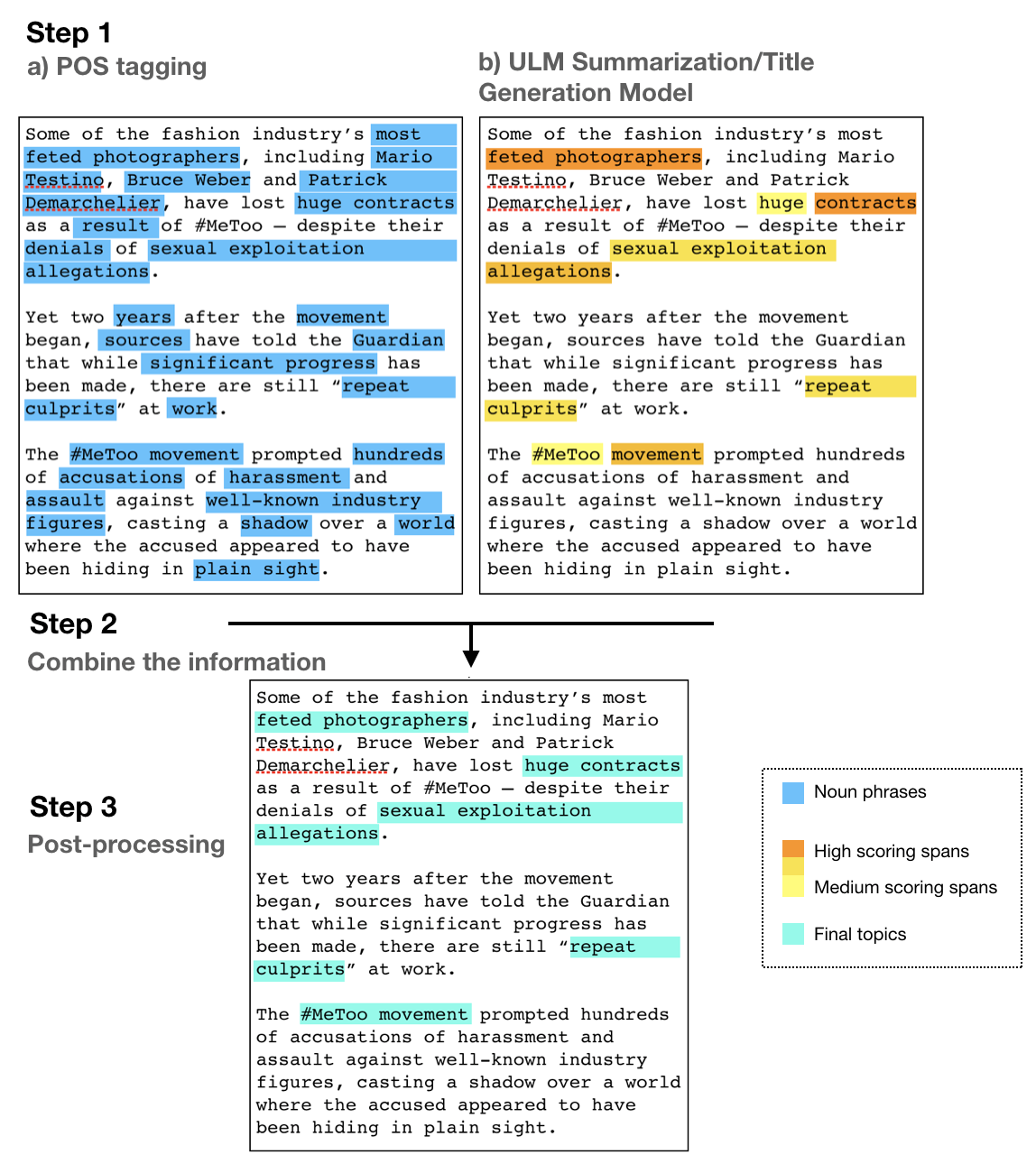}
    \caption{Topic identification. In Step 1 we generate (a) a list of noun phrases contained in the text as well as (b) a list of candidate spans obtained with a summarization/title generation model. In Step 2, we combine the information by using the quality of the overlap between noun phrases and candidate spans to generate a raw list of topics. In Step 3, some post-processing is applied to ensure basic deduplication.}
    \label{fig:strategy}
\end{figure}

In order to produce topics for a given text document, we obtain two lists of spans from a text. The first list is a wide list of all possible well-formed candidate phrases, disregarding their importance for the text. This list is purpose-oriented, in the sense that it depends on what kinds of topics we want to have in the final result. The second list is importance-oriented—it is a list of spans that may be not perfect phrases but have importance for the text. The details of the different steps of our strategy are shown in figure \ref{fig:strategy} .

We chose to generate the purpose-oriented list as a list of noun phrases using part of speech tagging provided by \textbf{}Natural Language Toolkit (nltk library). 
This could easily be switched out for another list of candidates, whether a list of noun phrases generated by a different natural language processing library, or a list of concepts of interest created for a specific purpose. For instance, we looked into using the open source library spaCy to identify noun chunks and obtained similar results.

Our novel approach in generating the importance-oriented list is in using a Universal Language Model trained for Summarization/Title Generation. 
The model generates a summary or a title by consequently selecting the most-suited words or spans. For use in topic generation, the only modification is that we output all highly ranked candidate spans, including those that did not make it to the final summary generation stage. Only the candidate spans that pass a quality threshold are retained as potential topics. We give more details on calculating this score below.

The two lists are then combined by prioritising the overlap between the spans, resulting in an intermediate list of topics. The overlap needs to meet certain criteria; in particular the noun in the noun phrase has to be present in the span overlap. 
 
The last step is to apply a simple post-processing to enhance the final quality, such as reducing duplicated information.

\subsection{Using a ULM to generate scores} \label{sec:ULM}

In order to obtain a list of spans scored and ranked by importance (see Figure \ref{fig:strategy} Step 1-b), we utilize the by-product of our document summarization/title generation model. Trained using a question-answer paradigm to generate news article titles, this model generates titles or bullet point summaries by using the text of a document in place of an external dictionary \cite{oleg2019headline}. 

The span selection is possible thanks to the model's prediction probabilities, represented by the logits of the start tokens and the end tokens of text spans. At each iteration, the start and end tokens with the highest logits are selected and used to generate the title span by span. When used for title generation or summarization, other spans are discarded. When we use this same model for keyphrase selection we investigate all candidate spans and assign them a score based on the sum of the logits of the first and last token.
\[score_\text{candidate span} = logit_\text{start token} + logit_\text{end token}\]
The candidate spans ($S_i$) are sorted by score and assigned a rank ($R_i$, the lower the value, the better the candidate) as well as a distance ($D_i$), with:

\[D_i = \frac{score_\text{top candidate} - score_i}{score_\text{top candidate}}\]

The list of candidate spans is obtained by selecting the top ranked candidates and applying a maximal distance threshold to ensure quality. More information about the distribution of these values is given in section \ref{sec:candidate span}. From that investigation, we show that the best results are obtained by picking all candidate spans that have rank lower than 15 and that are within a distance of 0.05 from the generation path. 

\subsection{Overlap between candidate spans and noun phrases}
We find all conventionally defined noun phrases in the text, and sort them by how much they overlap with our candidate spans. We define the overlap of a noun phrase with the candidate spans as a sum of lengths of all coinciding words divided by the length of the noun phrase. The pseudo-code is detailed in Figure \ref{fig:overlap}. We select the noun phrases that have highest overlaps with the spans from the importance-oriented list.

 \begin{figure}[h]
     \centering
     \begin{tabular}{l}
     \toprule
     Given a document text.\\  \\
     Find all noun phrases $noun\_phrases = [P_1, P_2, ..., P_n]$ in the text.\\ \\
     Produce a filtered list of candidate spans ($candidate\_spans$) by generation of a title: \\
     \hspace{.5cm}$candidate\_spans = [(S_1,D_1,R_1), ..., (S_n,D_n,R_n)]$,\\ 
     \hspace{.5cm}Each candidate-span $S_j$ has an associated rank $R_j$ and a distance $D_j$.\\
     \hspace{.5cm}Note that if $D_j$ is over a maximum quality threshold, $S_j$ is filtered out.\\ \\
     
     Initialize $overlapping\_phrases\_with\_candidate\_spans$ to an empty list\\
     \bf{for} $phrase$ in $noun\_phrases$:\\
     \hspace{.5cm}Initialize $length\_overlap = 0$\\
     \hspace{.5cm}Initialize $overlap\_contains\_noun = False$\\
     \hspace{.5cm}Initialize $overlapping\_candidate\_spans$ to an empty list\\
     \hspace{.5cm}\bf{for} $word$ in $phrase$:\\
     \hspace{.5cm}\hspace{.5cm}\bf{for} $candidate\_span$, $D$, $R$ in $candidate\_spans$:\\
     \hspace{.5cm}\hspace{.5cm}\hspace{.5cm}\bf{for} $word\_in\_span$ in $candidate\_span$:\\
     \hspace{.5cm}\hspace{.5cm}\hspace{.5cm}\hspace{.5cm}\bf{if} $word == word\_in\_span$:\\
     \hspace{.5cm}\hspace{.5cm}\hspace{.5cm}\hspace{.5cm}\hspace{.5cm}Increment $length\_overlap$ by $length(word)$\\
     \hspace{.5cm}\hspace{.5cm}\hspace{.5cm}\hspace{.5cm}\hspace{.5cm}Add $(candidate\_span,D,R)$ to $overlapping\_candidate\_spans$\\
     \hspace{.5cm}\hspace{.5cm}\hspace{.5cm}\hspace{.5cm}\hspace{.5cm}\bf{if} $\textnormal{POS}(word)$ is $\textnormal{noun}$:\\
     \hspace{.5cm}\hspace{.5cm}\hspace{.5cm}\hspace{.5cm}\hspace{.5cm}\hspace{.5cm}Set $overlap\_contains\_noun$ to $True$\\
     \hspace{.5cm}\bf{if} $overlap\_contains\_noun$ and $length\_overlap/length(phrase)>0.75$:\\
     \hspace{.5cm}\hspace{.5cm}Add ($phrase, overlapping\_candidate\_spans$) to $overlapping\_phrases\_with\_candidate\_spans$\\
     \bottomrule
     \end{tabular}
     \caption{Combining noun phrases with candidate spans generated by the ULM, using the overlap.}
     \label{fig:overlap}
 \end{figure}
 
\subsection{Post-processing topics}

The most crucial steps in the selection of topics was performed in the previous subsection. However the list obtained by checking the overlap between noun phrases and spans with significant summarization value can still have redundancy that is detrimental to the overall usefulness of the topics. We thus implement a simple post-processing step to further de-duplicate the topic labels and select the best.

We remove redundant information with a very basic de-duplication step, where we remove phrases with 50\% or more words contained in longer phrases. Other more sophisticated de-duplication methods could naturally be implemented.

For instance, with our simple approach the topics \textit{`Trump', `Kurdish forces', `Donald Trump', `Trump', `fighters'} and \textit{`US-backed Kurdish fighters'} would be de-duplicated to \textit{`Donald Trump' and `US-backed Kurdish fighters'}.

We also select higher value phrases by evaluating them against the 5 following dimensions:
\begin{enumerate}
    \item Distance criteria: we compute the mean distance across all the overlapping candidate spans corresponding to a given phrase. A smaller mean distance increases the value of the phrase. Mean distances over 0.4 do not change the value.
    \item Rank criteria: we compute the mean rank across all the overlapping candidate spans corresponding to a given phrase. Better mean rank (i.e. smaller numerical values) increase the overall value of the phrase. Mean ranks above 4 do not change the value.
    \item Number of candidate spans: a higher number of candidate spans increases the value of the phrase, and 4 or more spans gets the maximum value for this criteria.
    \item Number of words: a higher number of words in the phrase increases the value of that phrase. A phrase with 3 or more words gets the maximum value for this criteria. 
    \item Number of capitalized words: a higher number of words in the phrase starting with a capital letter increases the value of that phrase. If there are 3 or more capitalized words, the phrase gets the maximum value for this criteria. 
\end{enumerate}

The combination of those 2 steps allows us to proceed, for instance, from the highly repetitive and therefore less useful list of topics produced by the model:

\begin{center}\textit{`Qatar', `Palestinians', `West Bank', `ceasefire deal', `Qatar', `Qatar', `ceasefire deal', `Qatar', `Palestinians', `Israeli-occupied West Bank', `Qatar', `Qatar', `Qatar', `Palestinians', `Palestinians', `West Bank', `Qatar'}
\end{center}

to a de-duplicated list:

\begin{center}\textit{`ceasefire deal', `Qatar', `Palestinians', `Israeli-occupied West Bank' }
\end{center}

to	the final, re-ordered list:

\begin{center}\textit{`Israeli-occupied West Bank', `ceasefire deal', `Qatar'.}
\end{center}

This post-processing step allows us to have better quality topics, and could easily be tailored to choose the best topics for a given use-case.

\section{Inspecting candidate spans}\label{sec:candidate span}

\subsection{Generating all candidate spans}

In order the generate the candidate spans, we use a model trained on a task of title generation. This model has therefore never been exposed to topics before and is not in any way modified for generating topics.

At each step, the model finds the best span to include in the partial title it is tasked with generating. We track each step in the generation process through the span index. The model starts the generation at span index 0, getting the best span for this first iteration as well as all the other non-selected spans as a by-product. These are the candidate spans for span index 0. The model then searches for the next spans to include in the title, at span index 1, and repeats the process. Typically a title consists of 4-5 spans, but in rare cases the number of spans can reach up to 20 and more. 
As detailed in section \ref{sec:ULM}, the model can be used to give a score to each candidate-span (i.e. the sum of logits of its first and its last tokens). From the scores, we derive at each span index both the rank and the relative distance between a candidate span and the span actually selected for the title. 

\subsection{Properties of generated candidate spans}

The model was used to generate titles for 2000 randomly chosen English-language news documents published in May 2019. The information on candidate spans  is collected as a by-product. 

\begin{figure}
    \centering
    \includegraphics[width=12cm]{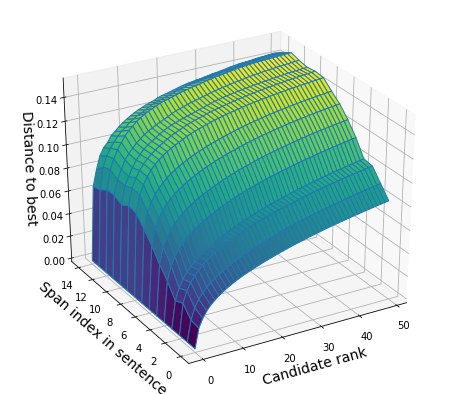}
    \caption{Distance of candidates to the best candidate. The distance depends on the span index in the generated title. The number of spans shown in the plot is limited to 15, and the number of candidates considered for each span is limited to 50. }
    \label{fig:span_candidate_distance_3D}
\end{figure}

In order to visualize the neighborhood of the generation path, we collected candidate spans from the 2000 generated titles. Figure \ref{fig:span_candidate_distance_3D} shows that information, truncated to 15 iterations of the title generation process (span index in sentence) and to the 50 best candidate spans (candidate rank).

Looking at the evolution of the distance along the "span index in sentence" axis, we observed that the shape of the distance curve is stabilized after approximately the 8th span index.

\begin{figure}
    \centering
    \includegraphics[width=10cm]{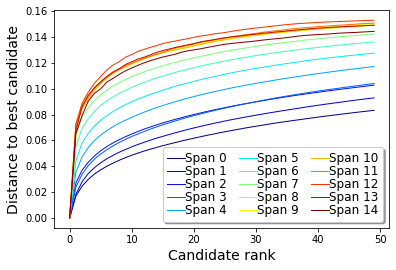}
    \caption{Distance of candidates to the best candidate. The distance depends on the span index in the generated title.}
    \label{fig:span_candidate_distance_2D}
\end{figure}

Looking at the evolution of the distance along the "candidate rank" axis, we see an initial sharp slope which then flattens out around ranks 10 to 20. This indicates a significant difference in quality between initial and late-phase candidate spans. We thus decided to not use the candidates with ranks higher than about 10 or 20 for the topic generation task. Figure \ref{fig:span_candidate_distance_2D}, a 2D projection along the axes considered, helps refine the choice of a reasonable threshold. 

The overall profile of the surface shown in Figure \ref{fig:span_candidate_distance_3D} clearly shows that there are a higher number of good quality candidate spans close to the best one (i.e. less than 0.04 in relative distance) when the model is first starting out on the generation path for the title (i.e. span index of 0) than when the model is trying to complete the title (i.e. span indices over 6). At that point, Figure \ref{fig:span_candidate_distance_2D} indicates that only perhaps the first five candidates are substantially better than the rest. This has an intuitive interpretation: It is easy to start a sentence—there are many valid ways to begin—but the options become increasingly constrained to maintain logical cohesion and fluency to the end of the sentence.

\subsection{Using the candidate spans as topics}

Our inspection of Figure \ref{fig:span_candidate_distance_3D} gives us a strong foundation to go about selecting an interesting subset of the scored candidate spans. We propose to use a threshold limit of 0.05 on the distance from the best candidate-span. We also propose to restrict ourselves to the candidate spans with a rank below 15. We therefore have as the model-produced importance list:
\[candidate\_spans = [(S_1,D_1,R_1), ..., (S_n,D_n,R_n)]\]
with $D_j <= 0.05$ and $R_j < 15$

As a side note, as we looked into candidate spans, we found that some of the filtered candidate spans already look like valid topics. Intuitively, this makes sense because the selected candidate spans almost made it into the title, and therefore reflect important concepts of the text. However, the title/summary generation aims for fluency, and some of the raw-candidate spans include verbs, articles, and punctuation that are not as useful for generating good topics. This is why producing the importance-oriented list is only one step in our topics generation process.

\section{Human evaluation}

\subsection{General setup}

In order to assess the quality of the generated topics we turn to human evaluation. We trained a group of 10 annotators to evaluate a list of topic labels associated with a document. We then contrasted the model's performance with an external data set of topics created by journalists for news articles. We found that the online news articles of The Huffington Post and The Guardian to be richly annotated with topic labels. We randomly selected 50 articles from The Guardian and 50 articles from Huffington Post, and collected the text and corresponding topics (referred to hereafter as "real topics"). For the same texts we generated topics by our method (referred to as "generated topics").

We set up the evaluation task by asking each annotator, hired through Odetta.ai \cite{odetta2019website}), to assess the quality of the topics on a 5-point scale: 

\begin{center}0 = VERY BAD, 1 = BAD, 2 = OK, 3 = GOOD or 4 = VERY GOOD.
\end{center}

The annotators work independently from each other and have access to only one article at a time. The text is displayed alongside the corresponding group of topic labels (the topics of the group are either all real or all generated) through the text annotation tool LightTag \cite{lighttag2019website}. Note that both the group containing the generated topics and the one with the real topics have about the same number of topics (most often between 3 and 5 topics). The annotators are not given any information about the origin of the topics they see. The order of real and generated examples is random.

\subsection{Evaluation with minimal instructions}

Before the labeling, the annotators were provided with minimal instructions, in order to avoid imposing a bias. The instructions are shown in Figure \ref{fig:instructions01}.

\begin{figure}[h]
\centering
\fbox{
\begin{minipage}{13 cm}
    The group of topics is given before the TEXT.
    Each topic in the group is separated by stars *
    
    The group has to be classified as VERY BAD or BAD or OK or GOOD or VERY GOOD.
    Criteria:
    \begin{itemize}
        \item Good group of topics covers the most important issues of the text.
        \item The topics in the group should be fluent and not too difficult to understand.
    \end{itemize}
    \caption{Initial instructions for evaluating topics.}
    \label{fig:instructions01}
\end{minipage}
}
\end{figure}

\begin{figure}
    \centering
    \includegraphics[width=14cm]{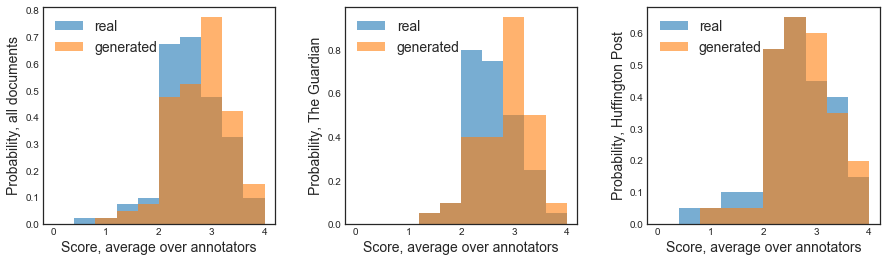}
    \caption{Distribution of scores assigned to topics. Left to right: (1) All 100 articles. (2) 50 articles from The Guardian. (3) 50 articles from Huffington Post.}
    \label{fig:distribution_scores_eval01_0to4}
\end{figure}

The distribution of the scores obtained in result is shown in Figure \ref{fig:distribution_scores_eval01_0to4}. For each article the scores were averaged over scores of all annotators.

The overall averages and medians of the scores are given in the table \ref{tab:evaluation01_values}.

\begin{table}
  \begin{center}
  \caption{First evaluation: scores of real and generated topics.}
  \label{tab:evaluation01_values}
    \begin{tabular}{l|l|S|S}
      \textbf{Documents} & \textbf{Real or generated} & \textbf{Average} & \textbf{Median}\\
      \hline
      \multirow{2}{*}{All} & Real & 2.63 & 2.67\\
      & Generated & 2.78 & 3.00\\ 
      \hline
      \multirow{2}{*}{The Guardian} & Real & 2.61 & 2.67\\
      & Generated & 2.83 & 3.00\\ 
      \hline
      \multirow{2}{*}{Huffington Post} & Real & 2.64 & 2.67\\
      & Generated & 2.73 & 2.67\\ 
      \hline
    \end{tabular}
  \end{center}
\end{table}   

\subsection{Evaluation with more guidance}

In order to reveal the influence of the instructions and to prompt annotators to apply more rigorous criteria, we modified our instructions as shown in the Figure \ref{fig:instructions02}, and also provided specific examples with suggested ranges of scores and descriptions of what the authors of this paper perceived as deficiencies in a set of topics.

\begin{figure}[h]
\centering
\fbox{
\begin{minipage}{13 cm}
    The group of topics is given before the TEXT.
    Each topic in the group is separated by stars *

    The group of topics has to be classified as VERY BAD or BAD or OK or GOOD or VERY GOOD. It is up to you to judge the group’s quality. 

    Some suggestions as to why the group of topics could be rated as lower quality:
    \begin{itemize}
        \item The topics do not cover the most important ideas of the text.
        \item The topics contain too much redundant information.
        \item Some topics in the group are difficult to understand or poorly worded.
        \item Some topics in the group describe minor details and do not add important information.
        \item Some topics are misleading or factually incorrect.
    \end{itemize}
    It is up to your judgement as to how much the score should be lowered in each case.
    
    \caption{Final instructions for evaluating topics.}
    \label{fig:instructions02}
\end{minipage}
}
\end{figure}

This evaluation used a new set of randomly selected 50 articles from The Guardian and 50 articles from The Huffington Post.

Having the list of possible deficiencies in the instructions, the annotators were now less generous with their scores. However, the overall preference for the generated topics persisted. The results from this second evaluation are provided in more detail below in the remaining part of this section.
 
\begin{figure}
    \centering
    \includegraphics[width=14cm]{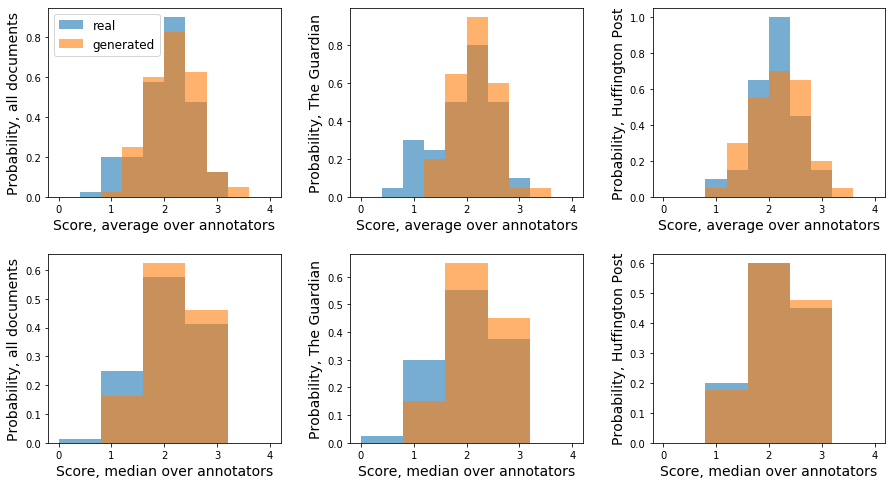}
    \caption{Distribution of scores assigned to topics. Top: for each article the scores are averaged over scores of all labelers. Bottom: for each article the median is taken over scores of all labelers. Left: All 100 articles. Middle: 50 articles from The Guardian. Right: 50 articles from Huffington Post.}
    \label{fig:distribution_scores_0to4}
\end{figure}

The distribution of the scores that annotators assigned to the real versus generated topics is shown in Figure \ref{fig:distribution_scores_0to4}.

The Table \ref{tab:evaluation_values} shows the main aggregation results of the evaluation.
\begin{table}
  \begin{center}
  \caption{Final evaluation: scores of real and generated topics.}
  \label{tab:evaluation_values}
    \begin{tabular}{l|l|S|S}
      \textbf{Documents} & \textbf{Real or generated} & \textbf{Average} & \textbf{Median}\\
      \hline
      \multirow{2}{*}{All} & Real & 2.07 & 2.10\\
      & Generated & 2.18 & 2.10\\ 
      \hline
      \multirow{2}{*}{The Guardian} & Real & 2.00 & 2.10\\
      & Generated & 2.16 & 2.20\\ 
      \hline
      \multirow{2}{*}{Huffington Post} & Real & 2.15 & 2.15\\
      & Generated & 2.20 & 2.10\\ 
      \hline
    \end{tabular}
  \end{center}
\end{table}  

In order to produce confidence estimates for the distribution of the gathered scores, we performed Bootstrap with 3 million samples (increasing the number of samples does not change our results). 

In the Bootstrap, each sample was obtained by two mutually independent random selections with replacement: selection of the 10 annotators and selection of the 100 articles (each article has two scores - one for real topics, another for generated topics). 

\begin{figure}
    \centering
    \includegraphics[width=16cm]{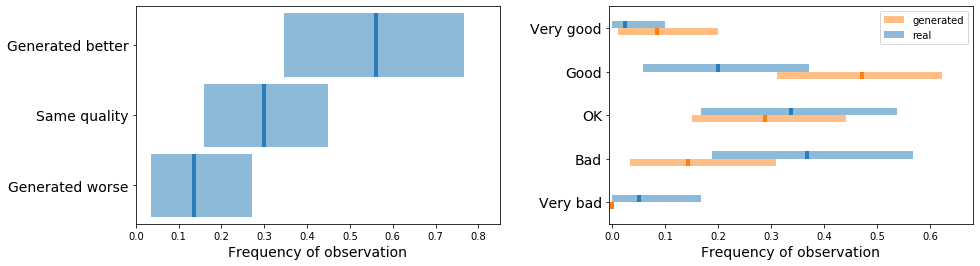}
    \caption{Bootstrap results.\\
    Left: Normalized distribution of the comparison between generated and real topics, with 95\% confidence intervals. The scores are given by the same person to the topics for the same texts.\\
    Right: Normalized distribution of the scores given by human labelers to generated and real topics, with 95\% confidence intervals shown.}
    \label{fig:bootstrap_results}
\end{figure}

The results of comparison of the scores given by the same annotator to the real versus generated topics of the same document are shown in Figure \ref{fig:bootstrap_results} (left), with 95\% confidence intervals. The distribution of the scores with 95\% confidence interval is shown in Figure \ref{fig:bootstrap_results} (right).

The Table \ref{tab:examples_scored} shows several examples of scored topics. 
The examples are chosen to represent the spectrum of the difference between the scores given to the generated vs real topics.

\begin{table}
 \caption{Examples of real and generated scored topics. The median scores are taken from all the labelers, with 0 the worst score and 4 the best. For instance, the underlying scores for the first row is [0, 0, 0, 0, 1, 1, 1, 1, 2, 2] for real topics and [1, 2, 2, 3, 3, 3, 3, 3, 3, 3] for generated topics.}
  \centering
  \begin{tabular}{p{6.8cm}p{8cm}}
    \toprule
    \begin{tabular}{p{1.2cm}p{3.8cm}p{1cm}}
        &
        \bf Topics & 
        \bf Median score 
    \end{tabular}
    & 
    \begin{tabular}{p{7.7cm}}
        \bf Title and text: only the top is shown here \\ \\
    \end{tabular}\\
    \toprule
    \begin{tabular}{r p{4cm} c}
        real & `Politics News', `Business', `Sports', `Technology', `Asia' & 1.0 \vspace{4pt} \\
        \midrule
        generated & `Hong Kong protests', `Houston Rockets official', `National Basketball Association', `Chinese Basketball Association', `China' & 3.0
    \end{tabular}
    &
    \begin{tabular}{p{8.9cm}}
    TITLE: NBA In Crisis After Bashing GM To Appease China\\
    TEXT: The National Basketball Association (NBA) came under fire on Monday for its response to a tweet by a Houston Rockets official in support of Hong Kong protests for democracy, the latest overseas business to run afoul of political issues in China. The Rockets’ general manager, Daryl Morey, apologized on Monday for the tweet he swiftly deleted on the weekend, but his support for the protests in the Chinese-ruled city angered Beijing, Chinese fans and the team’s partners in a key NBA market. ...
    \end{tabular} \\
    \bottomrule
    \begin{tabular}{r p{4cm} c}
        real & `Movies', `Margot Robbie', `Harley Quinn', `Dc Universe' & 2.0 \vspace{4pt} \\
        \midrule
        generated & `Harley Quinn', `Margot Robbie', `first trailer', `upcoming DC Universe film', `Jared Leto' & 3.0
    \end{tabular}
    &
    \begin{tabular}{p{8.9cm}}
    TITLE: Margot Robbie Returns As Harley Quinn In New 'Birds Of Prey' Trailer\\
    TEXT: Looks like Harley Quinn is coming back to the big screen, but this time without a Joker. That’s one of the things revealed in the first trailer for “Birds of Prey,” an upcoming DC Universe film centered around Margot Robbie’s popular villainess. Quinn is not a solo act. The trailer shows Robbie’s character joining up with female DC favorites like the Huntress, played by Mary Elizabeth Winstead, and Black Canary, played by Jurnee Smollett-Bell. ...
    \end{tabular} \\
    \bottomrule
    \begin{tabular}{r p{4cm} c}
        real & `Donald Trump', `Politics and Government', `2020 Election', `Joe Biden', `Chuck Schumer' & 2.0 \vspace{4pt}\\
        \midrule
        generated & `Trump administration', `Senate Intelligence Committees', `Republican-controlled Senate', `Senate Intelligence Committee', `GOP objections' & 2.0
    \end{tabular}
    &
    \begin{tabular}{p{8.9cm}}
    TITLE: Senate Unanimously Passes Measure Urging Release Of Ukraine Whistleblower Complaint\\
    TEXT: The Republican-controlled Senate on Tuesday unanimously passed a non-binding resolution calling on the Trump administration to release a complaint by a whistleblower from the intelligence community concerning President Donald Trump’s conduct with a foreign leader. The measure, introduced by Senate Minority Leader Chuck Schumer (D-N.Y.), expresses the sense of the Senate that the complaint, which the intelligence community’s inspector general deemed of “urgent concern,” ought to be provided to the House and Senate Intelligence Committees. ...
    \end{tabular} \\
    \bottomrule
    \begin{tabular}{r p{4cm} c}
        real & `Society And Culture', `Crime And Justice', `College Admissions In The United States', `Varsity Blues', `Gordon Caplan' & 2.0 \vspace{4pt}\\
        \midrule
        generated & `corporate law firm Willkie Farr', `New York', `largest college admissions', `Gordon Caplan', `former co-chairman' & 1.5
    \end{tabular}
    &
    \begin{tabular}{p{8.9cm}}
    TITLE: Lawyer Sentenced To One Month In Prison In U.S. College Admissions Scandal\\
    TEXT: The former co-chairman of the New York corporate law firm Willkie Farr \& Gallagher was sentenced on Thursday to one month in prison for his role in what prosecutors say is the largest college admissions scam uncovered in the United States. That is substantially less than the eight-month sentence federal prosecutors in Boston had sought for Gordon Caplan after he pleaded guilty to paying \textdollar75,000 to have a corrupt test proctor secretly correct his daughter’s answers on the ACT college entrance exam. ...
    \end{tabular} \\
    \bottomrule
    \begin{tabular}{r p{4cm} c}
        real & `Gavin Newsom', `Uber', `Lyft', `Governor Of California', `Independent Contractor' & 2.5 \vspace{4pt}\\
        \midrule
        generated & `many gig economy workers', `California Gov', `Gavin Newsom', `Assembly Bill', `California lawmakers' & 1.5
    \end{tabular}
    &
    \begin{tabular}{p{8.9cm}}
    TITLE: California Governor Signs Law That Could Upend Uber, Lyft\\
    TEXT: California Gov. Gavin Newsom (D) signed legislation on Wednesday that will reclassify many gig economy workers from independent contractors to employees, a change that could upend the business models of tech companies like Uber and Lyft. In a letter to California lawmakers on the bill signing, Newsom called Assembly Bill 5 “landmark legislation for workers and our economy. ...
    \end{tabular} \\
    \bottomrule
  \end{tabular}
  \label{tab:examples_scored}
\end{table}

\section{Conclusion}

In this paper we present a new approach for generating topics that requires neither distinct training data nor access to the entire corpus at inference time. We are able to to generate topic labels for a single document by utilizing a model trained for document summarization. The quality of the generated topics was deemed by double-blind trial to be on par with topic labels written by humans.

While utilizing the span candidates for generating topics and for some other usages, we were fascinated by the rich neighborhood of the generation path. The topics we generate are essentially concepts that 'wanted to be' in a title for the document but did not quite make the cut. For the purposes of the evaluation presented in this paper we used generation of a title restricted to reading the first several paragraphs of the text—as much length as allowed by the maximal input length of the standard BERT model. This is normally enough for getting all useful topics because, if the text is not too long, the most important topics are mentioned starting from the top of the text. As our evaluations with annotators show, this is indeed enough for typical news articles published by The Guardian and The Huffington Post.

For longer articles we use our summarization model which makes multiple runs, generating title-like sentences for each next chunk of text. In doing so, the generation picks up the most important concepts throughout the text. We can also change the criteria used for ranking. 

Finally, we have not discussed here the usage of our topics for clustering of documents, but a large fraction of our topics contain topics generic enough for this purpose. 

We are thankful to Delenn Chin, Vedant Dharnidharka and Wei Gong for reviewing the paper.

\bibliographystyle{unsrt}

\begin{thebibliography}{1}

\bibitem{eirini2019areview}
Eirini Papagiannopoulou, Grigorios Tsoumakas. 
\newblock A Review of Keyphrase Extraction.
\newblock {\em arXiv preprint arXiv:1905.05044v2}, 2019.

\bibitem{erion2019survey}
Erion Çano, Ondřej Bojar. 
\newblock Keyphrase Generation: A Multi-Aspect Survey.
\newblock {\em arXiv preprint arXiv:1910.05059}, 2019.

\bibitem{erion2019keyphrase}
Erion Çano, Ondřej Bojar. 
\newblock Keyphrase Generation: A Text Summarization Struggle.
\newblock {\em arXiv preprint arXiv:1904.00110v2}, 2019.

\bibitem{chang2009reading}
Jonathan Chang, Jordan Boyd-Graber, Sean Gerrish, Chong Wang, David M. Blei.
\newblock Reading Tea Leaves: How Humans Interpret Topic Models.
\newblock {Proceedings of the 22nd International Conference on Neural Information Processing Systems 22, pages 288–296}, 2009.

\bibitem{meng2017deepkeyphrase}
Rui Meng, Sanqiang Zhao, Shuguang Han, Daqing He, Peter Brusilovsky, Yu Chi.
\newblock Deep Keyphrase Generation.
\newblock {In Proceedings of the 55th Annual Meeting of the
Association for Computational Linguistics, pages 582–592}, 2017.

\bibitem{david2012probabilistic}
David M. Blei. 
\newblock Probabilistic topic models. 
\newblock {\em Communications of the ACM, 55(4):77–84}, 2012.

\bibitem{david2003latent}
David M. Blei, Andrew Y. Ng, and Michael I. Jordan. 
\newblock Latent dirichlet allocation.
\newblock {\em J. Mach. Learn. Res., 3:993–1022}, 2003.

\bibitem{akash2017autoencoding}
Akash Srivastava, Charles Sutton.
\newblock Autoencoding Variational Inference For Topic Models.
\newblock {\em arXiv preprint arXiv:1703.01488}, 2017.

\bibitem{Jason2019prediction}
Jason Ren, Russell Kunes, Finale Doshi-Velez.
\newblock Prediction Focused Topic Models via Vocab Selection.
\newblock {\em arXiv preprint arXiv:1910.05495}, 2019.

\bibitem{hasan2014survey}
Kazi Saidul Hasan and Vincent Ng.
\newblock Automatic Keyphrase Extraction: A Survey of the State of the Art.
\newblock {\em Proceedings of the 52nd Annual Meeting of the Association for Computational Linguistics (Volume 1: Long Papers)}, 2014.

\bibitem{liu2010topic}
Zhiyuan Liu, Wenyi Huang, Yabin Zheng, and Maosong Sun.
\newblock Automatic keyphrase extraction via topic decomposition. 
\newblock {\em In Proceedings of
the 2010 conference on empirical methods in natural language processing. Association for Computational Linguistics, pages 366–376}, 2010. 

\bibitem{bougouin2013topicrank}
Bougouin, A., Boudin, F. and Daille, B.
\newblock  TopicRank: Graph-based topic ranking for keyphrase extraction.
\newblock {\em Proceedings
of the 6th International Joint Conference on Natural Language Processing, IJCNLP}, 2013.

\bibitem{oleg2019headline}
Oleg Vasilyev, Tom Grek, John Bohannon.
\newblock Headline Generation: Learning from Decomposable Document Titles.
\newblock {\em arXiv preprint arXiv:1904.08455v3}, 2019.

\bibitem{odetta2019website}
\newblock https://odetta.ai/

\bibitem{lighttag2019website}
\newblock https://www.lighttag.io/

\end{thebibliography}

\end{document}